\documentclass{article}
\usepackage{ijcai20}

\usepackage{times}

\usepackage{soul}
\usepackage{url}
\usepackage[hidelinks]{hyperref}
\usepackage[utf8]{inputenc}
\usepackage[small]{caption}
\usepackage{graphicx}
\usepackage{amsmath}
\usepackage{booktabs}
\usepackage{mathrsfs}
\usepackage{multicol}
\usepackage{verbatim}
\usepackage{amsmath, amssymb,amsthm}
\usepackage{amsthm} % for proof environment
\usepackage{esvect}
\usepackage{bbm}
\usepackage{dsfont}
\usepackage{relsize}
\usepackage{adjustbox}
\usepackage{enumitem}
\usepackage{multirow}
\usepackage{graphicx}
\usepackage{algorithm}
\usepackage{algorithmic}
\usepackage{tabularx}
\usepackage{booktabs,subcaption,amsfonts,dcolumn}
\long\def\comment#1{}

\def\E{\mathbb{E}}

\title{Online Semi-Supervised Learning with Bandit Feedback}

\author{
Sohini Upadhyay, Mikhail Yurochkin, Mayank Agarwal, Yasaman Khazaeni and Djallel Bouneffouf\\
\affiliations
IBM Research\\
\emails
\{first.lastname\}@IBM.com
}

\begin{document}

\maketitle
\begin{abstract}
We formulate a new problem at the intersection of semi-supervised learning and contextual bandits, motivated by several applications including clinical trials and ad recommendations. We demonstrate how Graph Convolutional Network (GCN), a semi-supervised learning approach, can be adjusted to the new problem formulation. We also propose a variant of the linear contextual bandit with semi-supervised missing rewards imputation. We then take the best of both approaches to develop multi-GCN embedded contextual bandit. Our algorithms are verified on several real world datasets.
\end{abstract}

\section{Introduction}
We formulate the problem of Online Partially Rewarded (OPR) learning. Our problem is a synthesis of the challenges often considered in the semi-supervised and contextual bandit literature. Despite a broad range of practical cases, we are not aware of any prior work addressing each of the corresponding components. Next we justify each of the keywords and give motivating examples.
\begin{itemize}
    \item \textit{Online}: data is often naturally collected over time and systems are required to make predictions (take an action) before they are allowed to observe any response from the environment.
    \item \textit{Partially}: oftentimes there is no response available, e.g. a missing label or environment not responding to system's action.
    \item \textit{Rewarded}: in the context of online (multiclass) supervised learning we assume that the environment will provide the true label - however in many practical systems we can only hope to observe feedback indicating whether our prediction is good or bad (1 or 0 reward), the latter case obscuring the true label for learning.
\end{itemize}
Practical scenarios that fall under the umbrella of OPR range from clinical trials to dialog orchestration. In clinical trials, reward is partial, as patients may not return for followup evaluation. When patients do return, if feedback on their treatment is negative, the best treatment, or true label, remains unknown and the only available information is a reward of 0 for the treatment administered. In dialog systems, a user's query is often directed to a number of domain specific agents and the best response is returned. If the user provides negative feedback to the returned response, the best available response is uncertain and moreover, users can choose to not provide feedback at all. 

% multiclass semi-supervised learning where (3) we can only observe the true class when the correct prediction has been made. We observe that while this is a common real life scenario, only subsets of the problem are studied in different domains of Artificial Intelligence. First let us give some motivating examples: EXAMPLES. Next we review literature in two closely related domains and summarize it in the FANCY TABLE WITH CHECK MARKS.

In many applications, obtaining labeled data requires a human expert or expensive experimentation, while unlabeled data may be cheaply collected in abundance. Learning from unlabeled observations to improve prediction performance is the key challenge of semi-supervised learning \cite{chapelle2009semi}. One of the possible approaches is the continuity assumption, i.e. points closer to each other in the feature space are more likely to share a label \cite{seeger2000learning}. When the data has a graph structure, another approach is to perform node classification using graph Laplacian regularization, i.e. penalizing difference in outputs of the connected nodes \cite{zhu2003semi}. The latter approach can also be applied without the graph under the continuity assumption by building similarity based graph. We note that the problem of online semi-supervised leaning is rarely considered, with few exceptions \cite{Yver2009,valko2012online,Missing2020}. In our setting, the problem is further complicated by the bandit-like feedback in place of labels, rendering the existing semi-supervised approaches inapplicable. We will however demonstrate how one of the recent approaches, Graph Convolutional Networks (GCN) \cite{kipf2016semi}, can be extended to our setting.

The multi-armed bandit problem provides a solution to the exploration versus exploitation trade-off, informing a player how to pick within a finite set of decisions while maximizing cumulative reward in an online learning setting. Optimal solutions have been developed for a variety of problem formulations \cite{UCB,AuerCFS02,allesiardo2014neural,BouneffoufF16,BalakrishnanBMR18,Bouneffouf0SW19,LinBCR18,BouneffoufRC17,lin2020unified,BouneffoufRCF17}. These approaches do not take into account the relationship between context and reward, potentially inhibiting overall performance. In Linear Upper Confidence Bound (LINUCB) ~\cite{13,ChuLRS11,Survey2020,bouneffouf2020doublelinear,Corrupted2019} and in Contextual Thompson Sampling (CTS)~\cite{AgrawalG13}, the authors assume a linear dependency between the expected reward of an action and its context; the representation space is modeled using a set of linear predictors. These algorithms assume that the bandit can observe the reward at each iteration, which is not the case in our setting. Several authors have considered variations of partial/corrupted rewards \cite{bartok2014partial,gajanecorrupt}, however the case of entirely missing rewards has not been studied to the best of our knowledge.

% \cite{bartok2014partial} considered a kind of imperfect feedback called "Partial  Monitoring (PM)", developing a general framework for sequential decision making problems with imperfect feedback. This method assumes the availability of partial feedback, and does not consider the case of entirely missing feedback. \cite {gajanecorrupt} studied a variant of the stochastic multi-armed  bandit (MAB) problem in which the rewards are systematically corrupted motivated by privacy preservation in online recommender systems, which operates under the assumption that there exists a stochastic corruption process with known paramaters.

% \\QUESTION: Are we listing/summarizing our contributions explicitly anywhere? --- Yes

In this paper we focus on handling the problem of online semi-supervised learning with bandit feedback. We first review some existing methods in the respective domains and propose extensions to each of them to accommodate our problem setup. Then we proceed to combine the strengths of both approaches to arrive at an algorithm well suited for the Online Partially Rewarded learning as demonstrated with experiments on several real datasets. 

% In this work we consider natural extensions of both semi-supervised and MAB perspectives to solve (1-3). Specifically, we take one of the recent successes in semi-supervised learning - Graph Convolutional Network (GCN) and propose its extension to accommodate (1) and (3). Next we consider MAB problem formulation and suggest two imputation strategies for LINUCB - clustering based and one inspired by the GCN. We show theoretical guarantees for some of out proposed algorithms and compare them on multiple datasets.

\section{Preliminaries}
In this section we review two approaches coming from the respective domains of semi-supervised learning and contextual bandits, emphasizing their relevance and shortcomings in solving the OPR problem.

\subsection{Graph Convolutional Networks}
\label{subsec:gcn}
Neural networks have proven to be powerful feature learners when classical linear approaches fail. Classical neural network, Multi Layer Perceptron (MLP), is dramatically overparametrized and requires copious amounts of labeled data to learn. On the other hand, Convolutional Neural Networks are more effective in the image domain \cite{krizhevsky2012imagenet}, partially due to parameter sharing exploiting relationships between pixels. Image structure can be viewed as a grid graph where neighboring pixels are connected nodes. This perspective and the success of CNNs inspired the development of convolution on graphs neural networks \cite{henaff2015deep,defferrard2016convolutional,bronstein2017geometric} based on the concept of graph convolutions known in the signal processing communities \cite{shuman2013emerging}. 
Though all these works are in the realm of classical supervised learning, the idea of convolving signal over graph nodes is also widely applied in semi-supervised (node classification) learning \cite{belkin2006manifold}, where the graph describes relationships among observations (cf. grid graph of features (pixels) in CNNs). \cite{kipf2016semi} proposed Graph Convolutional Network (GCN), an elegant synthesis of convolution on graphs ideas and neural network feature learning capability, which significantly outperformed prior semi-supervised learning approaches on several citation networks and knowledge graph datasets.

To understand the GCN method, let $X \in \mathbb{R}^{T\times D}$ denote a data matrix with $T$ observations and $D$ features and let $A$ denote a positive, sparse, and symmetric adjacency matrix $A$ of size $T\times T$. The GCN embedding of the data with one hidden layer of size $L$ is $g(X) = \hat A \ \text{ReLU}(\hat A XW) \in \mathbb{R}^{T\times L}$, where $\hat A$ is degree normalized adjacency with self connections: $\hat A = (\mathcal{D}+\mathcal{I}_T)^{-1/2}(A + \mathcal{I}_T)(\mathcal{D} + \mathcal{I}_T)^{-1/2}$ and $\mathcal{D}_{ii} = \sum_{j=1}^T A_{ij}$ is the diagonal matrix of node degrees. $W\in \mathbb{R}^{D\times L}$ is a trainable weight vector. Resulting embedded data goes into the softmax layer and the loss for backpropagation is computed only on the \emph{labeled} observations. The product $\hat A X$ gives the one-hop convolution --- signal from a node is summed with signals from all of its neighbours achieving smooth transitions of the embeddings $g(X)$ over the data graph. Although a powerful semi-supervised approach, GCN is not suitable for the \textit{Online} and \textit{Rewarded} components of OPR. It additionally requires a graph as an input, which may not be available in some cases.

\subsection{Contextual Bandit}
Following \cite{langford2008epoch}, the contextual bandit problem is defined as follows.
At each time $t \in \{1,...,T\}$, a player is presented with a {\em context vector} $x_t\in \mathbb{R}^D$, $\|x_t\|\leq 1$   and must choose an arm $k \in \{ 1,...,K\}$.  $r_{t,k} \in [0,1]$ is the  reward of the action $k$ at time $t$, and $r_t \in [0,1]^K$ denotes a vector of  rewards for all arms at time $t$. 
We operate under the linear realizability assumption, i.e., there exist unknown weight vectors $\theta^*_k \in \mathbb{R}^D$ with $ \|\theta^*_k\|_2\leq 1$ for $k=1,\ldots,K$ so that
\begin{equation*}
\forall k, t: \; \E[r_{t,k} \vert x_t] = \theta_k^{*\top} x_t
%+ n_t.
\end{equation*}
Hence, the $r_{t,k}$ are independent random variables with expectation $x^\top_t \theta^*_k $.%, with $n_t$ some measurement noise.
%We also assume here that, the measurement noise $n_t$ is independent of everything and is $\sigma$-sub-Gaussian for some $\sigma >0$, i.e., $E[e^{\phi\, n_t} ] \leq \exp(\frac{\phi^2 \sigma^2}{2})$ for all $ \phi \in \mathbb{R}$.
%\begin{definition}[Cumulative regret]
%{The regret of an algorithm accumulated during $T$ iterations is given as:
%\begin{equation*}
%R(T) =\sum ^{T}_{t=1} r_{t,k^*(t)} - \sum^{T}_{t=1} r_{t,k(t)},
%\end{equation*}}
%\end{definition}

One solution to the contextual bandit problem is the LINUCB algorithm~\cite{13} where the key idea is to apply online ridge regression to incoming data to obtain an estimate of the coefficients $\theta_k$ for $k=1,\ldots,K$. At each step $t$, the LINUCB policy selects the arm with the highest upper confidence bound of the reward $k(t)= \text{argmax}_k (\mu_k + \sigma_k)$, where $\mu_k = \theta^{\top}_k x_{t}$ is the expected reward for arm $k$, $\sigma_k = \alpha \sqrt{x^{\top}_{t}\textbf{A}_k^{-1} x_{t}}$ is the standard deviation of the corresponding reward scaled by exploration-exploitation trade-of parameter $\alpha$ (chosen a priori) and $\textbf{A}_k$ is the covariance of the $k$-th arm context. LINUCB requires a reward for the chosen arm, $r_{t,k(t)}$, to be observed to perform its updates. In our setting reward may not be available at every step $t$, hence we need to adjust the LINUCB algorithm to learn from data with missing rewards.

\section{Proposed algorithms}
In this section we formally define Online Partially Rewarded (OPR) problem and present a series of algorithms, starting with natural modifications of GCN and LINUCB to suit the OPR problem setting and conclude with an algorithm building on strengths of both GCN and LINUCB.

\subsection{Problem setting}
We now formally define each of the OPR keywords:
\begin{itemize}
    \item \emph{Online}: at each step $t=1,\ldots,T$ we observe observation $x_t$ and seek to predict its label $\hat y_t$ using $x_t$ and possibly any information we had obtained prior to step $t$.
    \item \emph{Partially}: after we made the prediction $\hat y_t$, environment may not provide any feedback (we will use -1 to encode absence of feedback) and we have to proceed to step $t+1$ without knowledge of the true $y_t$.
    \item \emph{Rewarded}: suppose there are $K$ possible labels $y_t\in\{1,\ldots,K\}$. The environment at step $t$, if it responds to our prediction $\hat y_t$, will not provide true $y_t$, but instead a response $h_t \in \{-1,0,1\}$, where $h_t = 0$ indicates $\hat y_t \neq y_t$ and $h_t = 1$ indicates $\hat y_t = y_t$ (-1 indicates missing response).
\end{itemize}

\paragraph{Note on absence of environment response.} We assume that there is no dependence on $x_t$ in whether environment will respond or not. This is a common setting in semi-supervised learning \cite{chapelle2009semi} --- we have access to limited samples from the joint distribution of data and labels $\mathbb{P}(x,y)$ and larger amount of samples from the data marginal $\mathbb{P}(x)$ with the goal to infer $\mathbb{P}(y|x)$ using both. This assumptions is justified in some applications of interest, e.g. whether user will provide feedback to the dialog agent is independent of what the user asked.

\subsection{Rewarded Online GCN}
There are three challenges to be addressed to formulate Rewarded Online GCN (ROGCN): (i) online learning; (ii) the environment only responds with 0 or 1 to our predictions and (iii) the potential absence of graph information. As we shall see, there is a natural path from GCN to ROGCN. Suppose there is a small portion of data and labels available from the start, $X_0 \in \mathbb{R}^{T_0 \times D}$ and $y_0 \in \{-1,1,\ldots,K\}^{T_0}$, where $D$ is the number of features, $K$ is the number of classes and $T_0$ is the size of initially available data. When there is no graph available we can construct a $k$-NN graph ($k$ is a parameter chosen a priori) based on similarities between observations - this approach is common in convolutional neural networks on feature graphs \cite{henaff2015deep,defferrard2016convolutional} and we adopt it here for graph construction between observations $X_0$ to obtain graph adjacency $A_0$. We provide details in Section \ref{sec:experiments}. Now that we have $X_0,y_0,A_0$, we can train GCN with $L$ hidden units (a parameter chosen a priori) to obtain initial estimates of hidden layer weights $W_1 \in \mathbb{R}^{D \times L}$ and softmax weights $W_2 \in \mathbb{R}^{L \times K}$. Next we start to observe the stream of data --- as new observation $x_t$ arrives, we add it to the graph and data matrix, and append -1 (missing label) to $y$. Then we run additional training steps of GCN and output a prediction to obtain environment response $h_t \in \{-1,0,1\}$. Here 1 indicates correct prediction, hence we include it to the set of available labels for future predictions; 0 indicates wrong prediction and -1 an absence of a response, in the later two cases we continue to treat the label of $x_t$ as missing. This procedure is summarized in Algorithm \ref{alg:ROGCN}.

\begin{algorithm}[ht]
\caption{ROGCN}
 \label{alg:ROGCN}
 	\begin{footnotesize}
 \begin{algorithmic}[1]
   \STATE {\bfseries Input:} $W_1, W_2, X_0, y_0, \hat A_0$
   \STATE Set $X=X_0, y=y_0, \hat A = \hat A_0$
   \FOR{$t=T_0+1$ {\bfseries to} $T$}
   \STATE Append $x_t$ to $X$, -1 to $y$
   \STATE Update $\hat A$ with new edges if graph information is available or build $k$-NN similarity graph from $X$ to obtain $\hat A$
   \STATE Update $W_1$ and $W_2$ through GCN backpropagation with inputs $X,\hat A, y$
   \STATE Retrieve GCN prediction $\hat y_t$ and observe environment response $h_t\in \{-1,0,1\}$ for $\hat y_t$
   \IF{$h_t=1$}
   \STATE Replace last entry of $y$ with $\hat y_t$
   \ENDIF
   \ENDFOR
   \end{algorithmic}
   	\end{footnotesize}
\end{algorithm}

\subsection{Bounded Imputation LINUCB}
Contextual multi-armed bandits offer a powerful approach to online learning when true labels are not available and the environment's response to a prediction is observed at every observation instead. However, in our OPR problem setting, the environment may not respond to the agent for every observation. Classic bandit approach such as Linear Upper Confidence Bound (LINUCB) \cite{13} may be directly applied to OPR, however it would not be able to learn from observations without environment response. We propose to combine LINUCB with a user defined imputation mechanism for the reward when environment response is missing. In order to be robust to variations in the imputation quality, we only allow imputed reward to vary within agent's beliefs.
To make use of the context in the absence of the reward, we consider a user defined imputation mechanism $I(\cdot): \mathbb{R}^D \rightarrow \Delta^{K-1}$, which is expected to produce class probabilities for an input context $x_t$ to impute the missing reward. Typically any imputation mechanism will have an error of its own, hence we constrain the imputed reward to be within one standard deviation of the expected reward for the chosen arm:
\begin{equation}
\label{eq:impute_bounded}
r_t(k,x_t) = \max(\mu_k - \sigma_k, \min(I(x_t)_k, \mu_k + \sigma_k)).
\end{equation}
As in ROGCN, we can take advantage of small portion of data to initialize bandit parameters $b_k = \sum_{t: y_t=k}^{T_0} x_t$ and ${\bf{A}}_k = \sum_{t: y_t \neq -1}^{T_0} x_t x_t^{\top}$ for $k=1,\ldots,K$. Bounded Imputation LINUCB (BILINUCB) is summarized in Algorithm \ref{alg:BILINUCB}. %Later in the paper, we establish regret bounds for the BILINUCB, which are superior to LINUCB regret bounds.

\begin{algorithm}[ht]
\caption{BILINUCB}
 \label{alg:BILINUCB}
 	\begin{footnotesize}
 \begin{algorithmic}[1]
   \STATE {\bfseries Input:} $\alpha$, $b$, $\textbf{A}$, $I(\cdot)$
   \FOR{$t=T_0 + 1$ {\bfseries to} $T$}
   \STATE Update $I(\cdot)$ with $x_t$ and retrieve $I(x_t) \in \Delta^{K-1}$
   \FOR{all $k \in K$}
   \STATE $\theta_k \leftarrow \textbf{A}_k^{-1}*b_k$ \hspace{35pt} $\theta_k \leftarrow \theta_k/\|\theta_k\|_2$
   \STATE $\sigma_k \leftarrow$ $\alpha \sqrt{x^{\top}_{t}
   \textbf{A}_k^{-1} x_{t}}$ \hspace{15pt} $\mu_k \leftarrow \theta^{\top}_k x_{t}$
    \ENDFOR
%   \STATE Make a prediction $\hat y_t = \text{argmax}_{k\in K}  p_k$, and observe environment response $h_t\in \{-1,0,1\}$
    \STATE Predict $\hat y_t = \text{argmax}_{k}(\mu_k + \sigma_k)$, and observe environment response $h_t\in \{-1,0,1\}$
   \IF{ $h_t = 1$}
   \STATE $\textbf{A}_{k} \leftarrow \textbf{A}_{k} + x_{t} x^{\top}_{t}$ for $k=1,\ldots,K$
   \STATE $b_{\hat y_t} \leftarrow b_{\hat y_t} + x_{t}$
   \STATE Update $I(\cdot)$ with label $y_t=\hat y_t$
   \ELSIF{$h_t = 0$}
   \STATE $\textbf{A}_{\hat y_t} \leftarrow \textbf{A}_{\hat y_t} + x_{t}x^{\top}_{t}$
   \ELSIF{$h_t = -1$}
%   \STATE $r_t = p_{\hat y_t}\mathcal{I}_{I(x_t)_{\hat y_t}\geq p_{\hat y_t}} +$
%       $$+\max(\mu_{\hat y_t} - \sigma_{\hat y_t},I(x_t)_{\hat y_t})\mathcal{I}_{I(x_t)_{\hat y_t}<p_{\hat y_t}}$$
    \STATE $\textbf{A}_{\hat y_t} \leftarrow \textbf{A}_{\hat y_t} + x_{t}x^{\top}_{t}$
    \STATE $b_{\hat y_t} \leftarrow b_{\hat y_t} + r_t(\hat y_t, x_t) x_{t}$ (see Eq.~\eqref{eq:impute_bounded})
    \ENDIF
   \ENDFOR
   \end{algorithmic}
   	\end{footnotesize}
\end{algorithm}

\subsection{Multi-GCN embedded UCB}
We have presented two algorithms for OPR learning, however both approaches pose some limitations: ROGCN is unable to learn from missclassified observations and has to treat them as missing labels, while BILINUCB assumes linear relationship between data features and labels and even with perfect imputation is limited by the performance of the best linear classifier. Note that the bandit perspective allows one to learn from missclassfied observations, i.e. when the environment response $h_t=0$, and the neural network perspective facilitates learning better features such that linear classifier is sufficient. This observation motivates us to develop a more sophisticated synthesis of GCN and LINUCB approaches, where we can combine advantages of both perspectives.

To begin, we note that if $K=2$, a $h_t=0$ environment response identifies the correct class, hence the OPR reduces to online semi-supervised learning for which GCN can be trivially adjusted using ideas from ROGCN. To take advantage of this for $K>2$ we can consider a suite of GCNs for each of the classes, which then necessitates a procedure to decide which of the GCNs to use for prediction at each step.
% Our second observation is that the hidden layer of a GCN serves as a feature vector for softmax regression - a linear classifier. 
We propose to use a suite of class specific GCNs, where prediction is made using contextual bandit with context of $k$-th arm coming from the hidden layer representation of $k$-th class GCN and, when missing, reward is imputed from the corresponding GCN.

We now describe the multi-GCN embedded Upper Confidence Bound (GCNUCB) bandit in more details. Let $g(X)^{(k)} = \hat A \ \text{ReLU}(\hat A X W^{(k)}_1)$ denote the $k$-th GCN data embedding and let $g(X)^{(k)}_t$ denote the embedding of observation $x_t$. We will use this embedding (additionally normalized to unit $l_2$ norm) as context for the corresponding arm of the contextual bandit. The advantage of this embedding is its graph convolutional nature coupled with expressive power of neural networks. We note that as we add new observation $x_{t+1}$ to the graph and update weights of the GCNs, the embedding of the previously observed $x_1,\ldots,x_t$ evolves. Therefore instead of dynamically updating bandit parameters $b_k$ and $\textbf{A}_k$ as it was done in BILINUCB, we maintain set of indices for each of the arms $\mathcal{C}_k = \{t: \hat y_t = k \text{ or } h_t = 1\}$. At any step we can compute corresponding bandit context covariance and weight estimate using current embedding:
%  and history of rewards $\mathcal{R} = \{r_t \text{ for }t \in \mathcal{C}_k\}$, where $r_t=h_t$ if $h_t\neq -1$ or the $k$-th class probability according to $k$-th GCN if $h_t = -1$, i.e. imputed reward
%	\begin{footnotesize}
\begin{eqnarray}
& \textbf{A}_k = \sum_{t\in\mathcal{C}_k}g(X)^{(k)}_t {g(X)^{(k)}_t}^\top \label{eq:A_gcn} \\
& \theta_k = \textbf{A}_k^{-1} \sum_{t\in\mathcal{C}_k} r_{t,k} g(X)^{(k)}_t, \ \theta_k = \theta_k/\|\theta_k\|_2 \label{eq:theta_gcn}
\end{eqnarray}
%	\end{footnotesize}
 
where $r_{t,k}$ is the reward that was observed or imputed at step $t$ for arm $k$ (recall that we are imputing using prediction of the binary GCN corresponding to the arm chosen by the bandit). Now we can compute expected value and standard deviation for the reward on each arm. The prediction is made based on the upper confidence bounds for the rewards of the arms:
%	\begin{footnotesize}
\begin{equation}
\label{eq:arm_gcn}
    \begin{split}
    & \mu_k = \theta_k^\top g(X)^{(k)}_t \\
    & \sigma_k = \alpha \sqrt{{g(X)^{(k)}_t}^{\top}
   \textbf{A}_k^{-1} g(X)^{(k)}_t} \\
   & \hat y_t = \text{argmax}_{k} (\mu_k + \sigma_k).
    \end{split}
\end{equation}
%	\end{footnotesize}
Then we observe the environment response $h_t \in \{-1,0,1\}$. Unlike ROGCN, GCNUCB is able to learn from mistakes, i.e. when $h_t=0$ --- although as before we don't know the true class, we can be sure that it was not $\hat y_t$, hence we can use this information to improve GCN corresponding to the class $\hat y_t$. We summarize GCNUCB in Algorithm \ref{alg:GCNUCB}. Similarly to ROGCN and BILINUCB we can use a small amount of data $X_0$ and labels $y_0$ converted to binary labels $y^{(k)}_0 \in \{-1,0,1\}^{T_0}$ (as before -1 encodes missing label) for each class $k$ to initialize GCNs weights $W_1^{(k)},W_2^{(k)}$ for $k=1,\ldots,K$ and index sets $\mathcal{C}_k$ for each of the arms $k=1,\ldots,K$. Adjacency matrix if not given is obtained as in ROGCN.
\begin{algorithm}[ht]
\caption{GCNUCB}
 \label{alg:GCNUCB}
 	\begin{footnotesize}
 \begin{algorithmic}[1]
  \STATE {\bfseries Input:} $W^{(k)}_1, W^{(k)}_2, \mathcal{C}_k, r_{\cdot,k}, y^{(k)}_0 \ \forall k, \ X_0, \hat A_0, \alpha$
  \STATE Set $y^{(k)}=y^{(k)}_0 k=1,\ldots,K,\ X=X_0, \hat A = \hat A_0$
  \FOR{$t=T_0 + 1$ {\bfseries to} $T$}
  \STATE Append $x_t$ to $X$, -1 to each of $y^{(1)},\ldots,y^{(K)}$
  \STATE Update $\hat A$ with new edges if graph information is available or build  $k$-NN similarity graph from $X$ to obtain $\hat A$
  \STATE Update $W^{(k)}_1$ and $W^{(k)}_2$ through GCN backpropagation with inputs $X,\hat A, y^{(k)}$ for $k=1,\ldots,K$
  \STATE Retrieve embeddings $g(X)^{(k)}_t$ $\forall k$
  \STATE Compute $\textbf{A}_k$ (Eq. \eqref{eq:A_gcn}) and $\theta_k$ (Eq. \eqref{eq:theta_gcn}) $\forall k$
  \STATE Make prediction $\hat y_t$ using Eq. \eqref{eq:arm_gcn} and observe environment response $h_t$
  \IF{$h_t=1$}
  \STATE For each $k$ replace last entry of $y^{(k)}$ with 1 if $\hat y_t = k$ and 0 otherwise
  \STATE Append $t$ to each $\mathcal{C}_k$ and 1 to $r_{\cdot,k}$ if $\hat y_t = k$ and 0 otherwise
    \ELSIF{$h_t = 0$ (learning from mistakes)}
   \STATE Replace last entry of $y^{(\hat y_t)}$ with 0 
   \STATE Append $t$ to $\mathcal{C}_{\hat y_t}$ and 0 to $r_{\cdot,\hat y_t}$
   \ELSIF{$h_t = -1$ (imputing)}
    \STATE Append $t$ to $\mathcal{C}_{\hat y_t}$, output of $\hat y_t$-th GCN to $r_{\cdot,\hat y_t}$
  \ENDIF
  \ENDFOR
   \end{algorithmic}
   	\end{footnotesize}
\end{algorithm}

% \section{Algorithms analysis}
% \label{sec:analysis}

\section{Experiments}
\label{sec:experiments}

In this section we compare baseline method LINUCB which ignores the data with missing rewards to ROGCN, BILINUCB and GCNUCB --- algorithm proposed in this paper. We consider four different datasets: CNAE-9 and Internet Advertisements from the the UCI Machine Learning Repository\footnote{https://archive.ics.uci.edu/ml/datasets.html}, Cora \footnote{https://people.cs.umass.edu/~mccallum/data.html}, and Warfarin \cite{sharabiani2015revisiting}. Cora is naturally a graph structured data which can be utilized by ROGCN, BILINUCB with ROGCN based imputation and GCNUCB. For other datasets we use a 5-NN graph built online from the available data as follows.

Suppose at step $t$ we have observed data points $x_i \in \mathbb{R}^D$ for $i=1,\ldots,t$. Weights of the similarity graph computed as follows:
%	\begin{footnotesize}
\begin{equation}
\label{eq:knn}
A_{ij} = \exp\left(\frac{\|x_i - x_j\|^2_2}{\sigma^2}\right).
\end{equation}
%	\end{footnotesize}
As it was done by \cite{defferrard2016convolutional} we set $\sigma = \frac{1}{t}\sum_{i=1}^t d(i,i_k)$, where $d(i,i_k)$ denotes $L_2$ distance between observation $i$ and its $k$-th nearest neighbour indexed $i_k$. The k-NN adjacency $A$ is obtained by setting all but $k$ (excluding itself) corresponding closest entries of $A_{ij}$, $i,j=1,\ldots,t$ to 0 and symmetrizing. Then, as in \cite{kipf2016semi}, we add self connections and row normalize $\hat A = (\mathcal{D}+\mathcal{I}_T)^{-1/2}(A + \mathcal{I}_T)(\mathcal{D} + \mathcal{I}_T)^{-1/2}$, where $\mathcal{D}_{ii} = \sum_{j=1}^T A_{ij}$ is the diagonal matrix of node degrees.

For pre-processing we discarded features with large magnitudes (3 features in Internet Advertisements and 2 features in Warfarin) and row normalized all observations to have unit $l_1$ norm.

For all of our algorithms that use GCN we use default parameters of the GCN and Adam optimizer \cite{kingma2014adam}. Default parameters are as follows: 16 hidden units, learning rate of 0.01, 0.0005 weight decay, and dropout of 0.5.

\begin{table*}[ht]
\centering
\caption{Total average accuracy}
\vskip 7pt
\label{table:accuracy}
%\begin{subtable}[h]{0.32\linewidth}\centering
\scalebox{0.81}{
\begin{tabular}{p{0.18\linewidth}|p{0.13\linewidth}|p{0.13\linewidth}|p{0.13\linewidth}|p{0.13\linewidth}}
\multicolumn{5}{c}{25\% Missing labels}
\\ \toprule
        & CNAE-9    & Internet Ads      & Warfarin      & Cora             \\ \midrule
LINUCB   & 67.57 $\pm$ 2.90 & 90.08 $\pm$ 0.64 & 53.70 $\pm$ 0.77 & 38.06 $\pm$ 3.45  \\ 
ROGCN & 64.73 $\pm$ 2.67 & 88.22 $\pm$ 1.73 & 47.72 $\pm$ 9.40 & 48.57 $\pm$ 7.75  \\ 
BILINUCB-GCN & 67.27 $\pm$ 2.79 & 89.91 $\pm$ 0.73 & 53.70 $\pm$ 0.77 & 37.66 $\pm$ 3.92   \\ 
BILINUCB-KMeans & 67.69 $\pm$ 4.30 & 90.37 $\pm$ 0.63 & 52.53 $\pm$ 4.83 & 39.11 $\pm$ 2.68   \\
GCNUCB & \textbf{77.10 $\pm$ 1.89} & \textbf{93.14 $\pm$ 0.39} & \textbf{55.19 $\pm$ 3.40} & \textbf{66.01 $\pm$ 1.35}   \\ \bottomrule
\end{tabular}
}
\\
\vspace{\baselineskip}
\scalebox{0.81}{
\begin{tabular}{p{0.18\linewidth}|p{0.13\linewidth}|p{0.13\linewidth}|p{0.13\linewidth}|p{0.13\linewidth}}
\multicolumn{5}{c}{50\% Missing labels}
\\ \toprule
        & CNAE-9    & Internet Ads      & Warfarin      & Cora             \\ \midrule
LINUCB   & 64.25 $\pm$ 3.55 & 88.62 $\pm$ 0.67 & 51.87 $\pm$ 5.12 & 38.85 $\pm$ 2.74 \\ 
ROGCN & 65.96 $\pm$ 3.69 & 88.38 $\pm$ 1.93 & 49.37 $\pm$ 8.29 & 47.71 $\pm$ 9.25  \\ 
BILINUCB-GCN & 63.52 $\pm$ 3.31 & 88.40 $\pm$ 0.73 & 51.75 $\pm$ 5.32 & 38.08 $\pm$ 2.97   \\ 
BILINUCB-KMeans & 67.37 $\pm$ 5.18 & 89.95 $\pm$ 0.66 & 54.20 $\pm$ 0.30 & 39.20 $\pm$ 1.76   \\ 
GCNUCB & \textbf{74.55 $\pm$ 1.82}  & \textbf{92.62 $\pm$ 0.37} & \textbf{56.51 $\pm$ 3.43} & \textbf{63.47 $\pm$ 2.26}   \\ \bottomrule
\end{tabular}
}
\vspace{\baselineskip}
\scalebox{0.81}{
\begin{tabular}{p{0.18\linewidth}|p{0.13\linewidth}|p{0.13\linewidth}|p{0.13\linewidth}|p{0.13\linewidth}}
\multicolumn{5}{c}{75\% Missing labels}      
\\ \toprule
        & CNAE-9    & Internet Ads      & Warfarin      & Cora             \\ \midrule
LINUCB   & 61.67 $\pm$ 3.16 & 86.66 $\pm$ 0.99 & 52.99 $\pm$ 2.61& 33.92 $\pm$ 0.04  \\ 
ROGCN & 65.67 $\pm$ 5.28 & 88.31 $\pm$ 1.81 & 47.48 $\pm$ 5.41 & 49.63 $\pm$ 5.06  \\ 
BILINUCB-GCN & 61.36 $\pm$ 3.79 & 86.68 $\pm$ 1.04 & 50.04 $\pm$ 11.44 & 32.21 $\pm$ 5.99   \\ 
BILINUCB-KMeans & 57.16 $\pm$ 3.57 & 88.21 $\pm$ 0.99 & 51.21 $\pm$ 7.12 & 32.51 $\pm$ 4.98   \\ 
GCNUCB & \textbf{70.82 $\pm$ 2.33} & \textbf{91.45 $\pm$ 0.89} & \textbf{53.31 $\pm$ 2.98}& \textbf{58.29 $\pm$ 2.80} \\ \bottomrule
\end{tabular}}\\
%\end{subtable}
\end{table*}

% \begin{table}[h]
% 	\centering
% 	\caption{Datasets}
% 	\label{table:Datasets}
% 	\resizebox{\columnwidth}{!}{
% 		\begin{tabular}{l|l|l|l}

% Datasets                & Instances & Features & Classes \\ \hline
% CNAE-9                  & 1080      & 856      & 9       \\
% Internet Advertisements & 3279      & 1555     & 2       \\
% Cora                    & 2708      & 1433     & 7        \\
% Warfarin                & 5528      & 91       & 3      
% \end{tabular}
% 	}
% 	\label{table:Synthetic}
% \end{table}
% \subsection{Datasets}
% We compare the LINUCB, ROGCN, BILINUCB, and GCNUCB approaches to the OPR problem on five different datasets: CNAE-9 and Internet Advertisements from the the UCI Machine Learning Repository\footnote{https://archive.ics.uci.edu/ml/datasets.html}, Cora \cite{kipf2016semi}, and Warfarin \cite{sharabiani2015revisiting}. Cora has a natural adjacency matrix that we use in ROGCN, BILINUCB, and GCNUCB instead of generating an adjacency matrix with online $k$-NN as described in the algorithm formulations and supplement. We discard features and row normalize as detailed in the Supplement. The final details of each dataset are summarized in Table \ref{table:Datasets}. 

To emulate the OPR setting we randomly permute the order of the observations in a dataset and remove labels for some portion (we experiment with three settings: 25\%, 50\% and 75\% missing labels) of the observations chosen at random. For all methods we consider initial data $X_0$ and $y_0$ to represent a single observation per class chosen randomly ($T_0=K$). At a step $t=T_0+1,\ldots,T$ each algorithm is given a feature vector $x_t$ and is ought to make a prediction $\hat y_t$. The environment response $h_t \in \{-1,0,1\}$ is then observed and algorithms moves onto step $t+1$. To compare performance of different algorithms at each step $t$ we compare $\hat y_t$ to true label $y_t$ available from the dataset (but concealed from the algorithms themselves) to evaluate running accuracy. Defined as such, accuracy is inversely proportional to regret.

\paragraph{Imputation Methods.} We test two different imputation functions $I(\cdot)$ for BILINUCB - a ROGCN and simple k-means clustering with 10 clusters. Henceforth, we denote these two approaches as BILINUCB-GCN and BILINUCB-KMeans. In BILINUCB-GCN we update ROGCN with incoming observations and use the softmax class prediction to impute missing reward when needed. In BILINUCB-KMeans, we use the mini-batch k-means algorithm to cluster incoming observations online and impute missing reward with the average non-missing reward of all observations in the corresponding cluster.

\begin{figure}[h]
%\vspace{-0.15in}
\centering
\begin{subfigure}{0.48\linewidth}
  \centering
  \includegraphics[width=\linewidth]{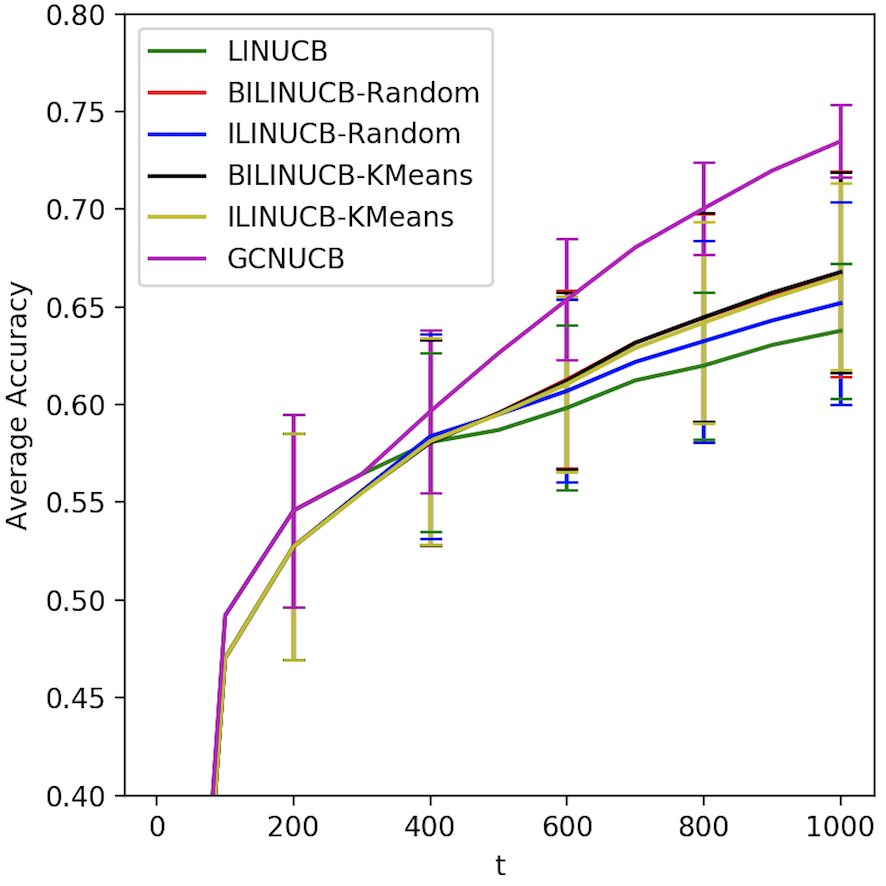}
  \caption{Accuracy on CNAE-9;\\ 50\% missing labels}
  \label{CNAE90.5M}
\end{subfigure}%
\begin{subfigure}{0.48\linewidth}
  \centering
  \includegraphics[width=\linewidth]{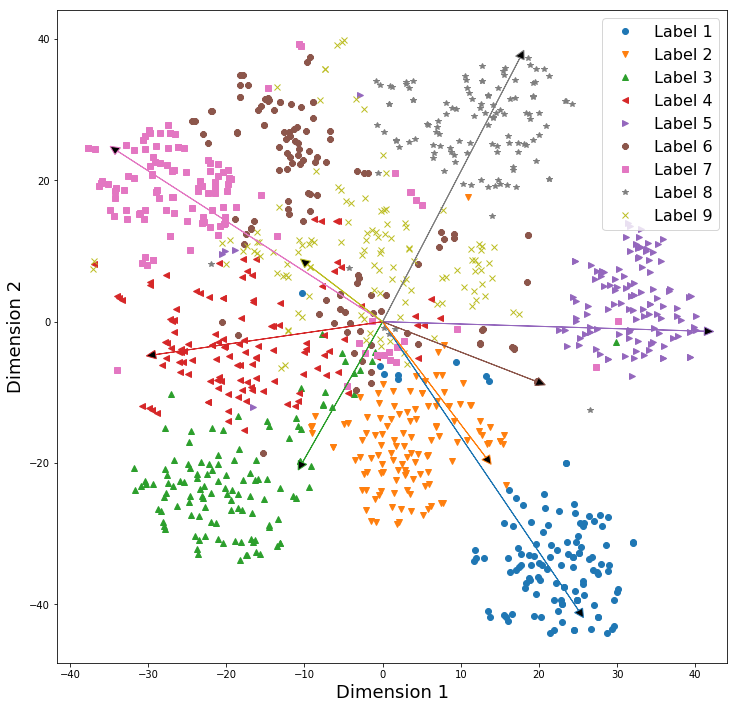}
  \caption{t-SNE embeddings of context and bandit weight vectors for LINUCB}
  \label{LINUCB}
\end{subfigure}
\begin{subfigure}{0.48\linewidth}
  \centering
  \includegraphics[width=\linewidth]{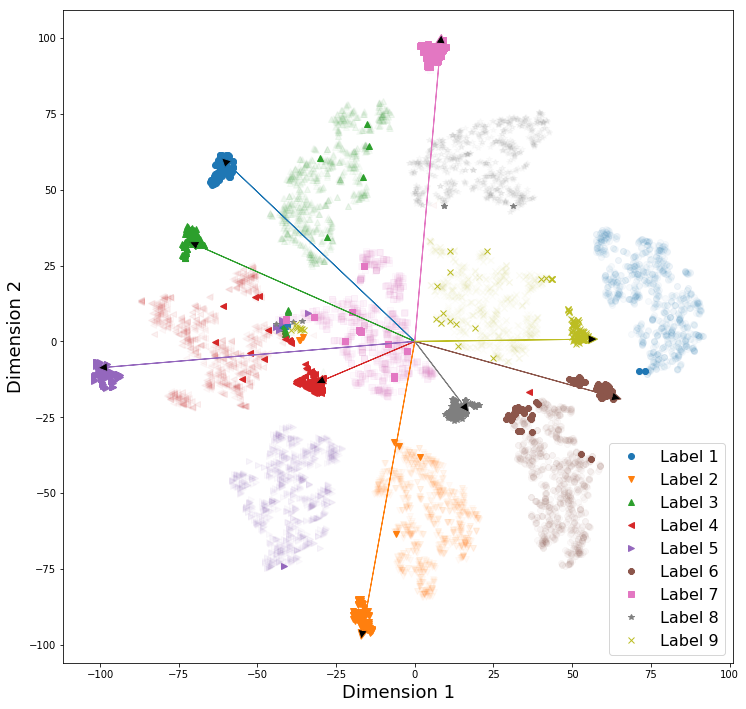}
  \caption{t-SNE embeddings of context and bandit weight vectors for GCNUCB}
  \label{GCNUCB}
\end{subfigure}
%\caption{Total Average Reward for Customer Assistant}
%\vspace{-0.15in}
\end{figure}

\paragraph{Running accuracy results.} We noticed that BILINUCB with both imputation approaches and GCNUCB are more robust to data ordering when we use baseline LINUCB for first 300 steps and then proceed with the corresponding algorithm (see Figure \ref{CNAE90.5M} where aforementioned algorithms and LINUCB perform the same until step 300 and then have individual running accuracies). For all LINUCB based approaches we used exploration-exploitation trade-off parameter $\alpha=0.25$. Results are summarized in Table \ref{table:accuracy}. Since ordering of the data can affect the problem difficulty, we performed 10 data resampling for each setting to obtain error bars.

% **MISHA EXPLAIN DELAYED START****For BILINUCB-GCN, BILINUCB-KMeans, and GCNUCB we use a delayed start parameter, $d$, where we run LINUCB for $d$ events to better initialize bandit parameters before starting BILINUCB-GCN, BILINUCB-KMeans, or GCNUCB. We use $\alpha=0.25$ and $d=300$ for LINUCB, BILINUCB-GCN, BILINUCB-KMeans, and GCNUCB, and a learning rate of 0.01, L2 regularization coefficient of 0.0005, 16 dimensional hidden layer, and dropout of 0.5 for ROGCN, BILINUCB-GCN, BILINUCB-KMeans, and GCNUCB. Though the true class is concealed from the algorithms themselves, for the purposes of evaluation we utilize an accuracy metric that checks the equality of the selected class and the true class. Defined as such, accuracy is inversely proportional to regret. In the following experiments, all of the algorithms are initialized with $K$ events with one correct decision for each of the $K$ classes, but are only evaluated on subsequent events.  We randomly simulate 25\%, 50\%, and 75\% of reward as missing for every dataset and report the final accuracy of each method in Table \ref{table:accuracy}.

GCNUCB outperforms the LINUCB baseline and our other proposed methods in all of the experiments, validating our intuition that a method synthesizing the exploration capabilities of bandits coupled with the effective feature representation power of neural networks is the best solution to the OPR problem. We see the greatest increase in accuracy between GCNUCB and the alternative approaches on the Cora dataset which has a natural adjacency matrix. This suggests that GCNUCB has a particular edge in OPR applications with graph structure. Such problems are ubiquitous. Consider our motivating example of dialog systems - for dialog systems deployed in social network or workplace environments, there exists graph structure between users, and user information can be considered alongside queries for personalization of responses. 

\paragraph{Role of bounding the imputed reward.} Notice that on average, a BILINUCB method outperforms LINUCB and ROGCN. To understand the role of these imputation bounds, we analyze the effects of random imputation. We denote this use of random imputation as BILINUCB-Random and ILINUCB-Random as the same without bounding the imputed reward. We define BILINUCB-KMeans and ILINUCB-KMeans similarly. We summarize the overall accuracy of each method on CNAE-9 in Table \ref{table:Bounds}.

As the purpose of these bounds is to correct for errors in the imputation method, we expect to see its impact the most when imputation is inaccurate. This is exactly what we see in Table \ref{table:Bounds} and Figure \ref{CNAE90.5M}. When we use a reasonable imputation method, k-means, the imputation bounds do not improve, or only make slight improvements to the results. The improvement gain is much more evident with random imputation, and across both imputation methods, the bounds have a larger impact when there is more reward missing.   

\paragraph{Visualizing GCNUCB context space.} Recall that the context for each arm of GCNUCB is provided by the corresponding binary GCN hidden layer. The motivation for using binary GCNs to provide the context to LINUCB is the ability of GCN to construct more powerful features using graph convolution and neural networks expressiveness. To see how this procedure improves upon the baseline LINUCB utilizing input features as context, we project the context and the corresponding bandit weight vectors, $\theta_1,\ldots,\theta_K$, for both LINUCB and GCNUCB to a 2-dimensional space using t-SNE \cite{maaten2008visualizing}. In this experiment we analyzed CNAE-9 dataset with 25\% missing labels. Recall that the bandit makes prediction based on the upper confidence bound of the regret: $\text{argmax}_k(\theta_k^\top x_{k,t} + \sigma_k)$ and that $x_{k,t}=x_t \ \forall k=1,\ldots,K$ for LINUCB and $x_{k,t}=g(X)^{(k)}_t$ for GCNUCB. To better visualize the quality of the learned weight vectors, for this experiment we set $\alpha=0$ and hence $\sigma_k=0$ resulting in a greedy bandit, always selecting an arm maximizing expected reward $\theta_k^\top x_{t,k}$. In this case, a good combination of contexts and weight vectors is the one where observations belonging to the same class are well clustered and corresponding bandit weight vector is directed at this cluster.
For LINUCB (Figure \ref{LINUCB}, 68\% accuracy) the bandit weight vectors mostly point in the direction of their respective context clusters, however the clusters themselves are scattered, thereby inhibiting the capability of LINUCB to effectively distinguish between different arms given the context.
In the case of GCNUCB (Figure \ref{GCNUCB}, 77\% accuracy) the context learned by each GCN is tightly clustered into two distinguished regions - one with context for corresponding label and binary GCN when it is the correct label (points with bolded colors), and the other region with context for the label and GCN when a different label is correct (points with faded colors). The tighter clustered contexts allow GCNUCB to effectively distinguish between different arms by assigning higher expected reward to contexts from the correct binary GCN than others, thereby resulting in better performance of GCNUCB than other methods.

\begin{table}[ht]
\centering
\caption{CNAE-9 total average accuracy}
\vskip 7pt
\label{table:Bounds}{}
\resizebox{\columnwidth}{!}{
\begin{tabular}{l|l|l}
\toprule
\% Reward Missing & ILINUCB-Random   & BILINUCB-Random           \\ \midrule
25                 & 67.29 $\pm$ 4.15 & \textbf{67.65 $\pm$ 4.30} \\ 
50                 & 65.67 $\pm$ 5.20 & \textbf{67.19 $\pm$ 5.37} \\ 
75                 & 49.77 $\pm$ 4.68 & \textbf{56.36 $\pm$ 3.71} \\
\end{tabular}}
\resizebox{\columnwidth}{!}{
\vspace{\baselineskip}
\begin{tabular}{l|l|l}
\toprule
\% Reward Missing & ILINUCB-KMeans            & BILINUCB-KMeans           \\ \midrule
25                 & \textbf{67.92 $\pm$ 3.98} & 67.69 $\pm$ 4.30          \\ 
50                 & 67.14 $\pm$ 4.84          & \textbf{67.37 $\pm$ 5.18} \\ 
75                 & 56.62 $\pm$ 4.40          & \textbf{57.16 $\pm$ 3.57} \\
\bottomrule
\end{tabular}
}
\end{table}

\section{Conclusion and Discussion}
We have defined and studied the problem of Online Partially Rewarded (OPR) learning, which combines challenges from semi-supervised learning and multi-armed contextual bandits. We have developed ROGCN and BILINUCB - extensions of popular algorithms in the corresponding domains to solve the OPR problem. Our main contribution, GCNUCB algorithm, is the efficient synthesis of the strengths of the two approaches. Our experiments show that GCNUCB, which combines feature extraction capability of the graph convolution neural networks and natural ability of contextual bandits to handle online learning with reward (instead of labels), is the best approach for OPR across a LINUCB baseline and other algorithms that we proposed. %Next we discuss several limitations of GCNUCB and how they could be addressed in the future work. 
%\paragraph{Memory limitation} In our current implementation of GCNUCB we use all of the data seen so far to update parameters of binary GCNs. This may not be a permissible choice for large data sizes, however some of the recent work \cite{hamilton2017inductive,chen2018fastgcn} has already proposed variants of the mini-batch GCN training. We think adopting this line of work in the OPR context will help to make GCNUCB applicable to larger datasets.

%\paragraph{Understanding of graph embeddings as context}We acknowledge that satisfactory theoretical analysis of GCNUCB, as in most cases where neural networks are used, may not be achievable at the moment, however we think that analyzing graph convolutional embeddings as bandit context, considering simpler approaches such as polynomial graph convolutional filters \cite{sandryhaila2013discrete}, is an important direction for joint studies of semi-supervised and online learning with bandit feedback.

\bibliographystyle{named}
\interlinepenalty=10000
\bibliography{ijcai20}

\end{document}